\documentclass[lettersize,journal]{IEEEtran}
\usepackage{amsmath,amsfonts}
\usepackage{algorithmic}
\usepackage{algorithm}
\usepackage{array}
\usepackage[caption=false,font=normalsize,labelfont=sf,textfont=sf]{subfig}
\usepackage{textcomp}
\usepackage{stfloats}
\usepackage{url}
\usepackage{verbatim}
\usepackage{graphicx}
\usepackage{cite}
\usepackage{multirow}
\usepackage{booktabs}
\usepackage{xcolor}
\begin{document}

\title{DINOFAS: Bonafide-Centric Representation Learning for Face Anti-Spoofing}

\author{Shakeel Muhammad Ibrahim, Muhammad Sohail Ibrahim, Muhammad Shahid Jabbar, and Shujaat Khan \thanks{S. M. Ibrahim is with Geomexsoft Co. Ltd., Chuncheon-si, Republic of Korea.} \thanks{M.S. Jabbar and S. Khan are with SDAIA-KFUPM Joint Research Center for Artificial Intelligence (JRC-AI), King Fahd University of Petroleum \& Minerals, Dhahran, 31261, Saudi Arabia.} \thanks{M.S. Ibrahim is with Interdisciplinary Research Center for Intelligent Secure Systems (IRC-ISS), King Fahd University of Petroleum \& Minerals, Dhahran, 31261, Saudi Arabia.} \thanks{S. Khan is also with Department of Computer Engineering, College of Computing and Mathematics,~King Fahd University of Petroleum \& Minerals, Dhahran, 31261, Saudi Arabia.}}

\markboth{Journal of \LaTeX\ Class Files,~Vol.~14, No.~8, August~2021}%
{Shell \MakeLowercase{\textit{et al.}}: A Sample Article Using IEEEtran.cls for IEEE Journals}


\maketitle

\begin{abstract}
Face anti-spoofing (FAS) models often learn attack- and dataset-specific cues that degrade performance under unseen acquisition conditions and presentation attacks. We present DINOFAS, a DINOv3-based framework that formulates FAS as bonafide-centric representation learning. The proposed framework combines parameter-efficient DINOv3 adaptation using low-rank adaptation (LoRA), an attention style-content disentanglement (ASCD) module for reducing domain-specific appearance variations, and a dual-branch architecture that separates prototype-based liveness learning from attack-semantic supervision. The model is jointly optimized using a bonafide-centric prototype objective, an attack-semantic classification loss applied to detached features, and orthogonality regularization. DINOFAS achieves an average HTER of 9.09\% under the standard OCIM leave-one-domain-out protocol, where training is performed on three source datasets and evaluation is conducted on the held-out unseen domain. We further evaluate synthetic-to-real generalization by training exclusively on SynthASpoof and testing on four unseen real-world datasets, where DINOFAS achieves the lowest average HTER of 21.41\% and the highest average AUC of 86.14\% among the compared methods. Evaluations on OULU-NPU and SiW-Mv2 additionally demonstrate robustness to variations in acquisition conditions, devices, and unseen presentation attacks.
\end{abstract}


\section{Introduction}
\IEEEPARstart{F}{ace} recognition (FR) systems have become an integral component of modern biometric authentication, enabling secure access to mobile devices, financial services, border control, and identity verification. Their remarkable progress over the past decade has largely been driven by advances in deep learning and large-scale training datasets \cite{george2024edgeface}. Despite these advances, FR systems remain vulnerable to presentation attacks (PAs), in which an adversary attempts to impersonate a legitimate user using printed photographs, replayed videos, or three-dimensional masks \cite{abdullakutty2022deep, chen2021dual}. Successful presentation attacks can compromise the security of biometric systems, leading to identity theft, unauthorized access, and financial fraud. Consequently, face anti-spoofing (FAS) has become an indispensable component of practical biometric systems.

\begin{figure*}[]
    \centering
    \includegraphics[width=0.85\linewidth]{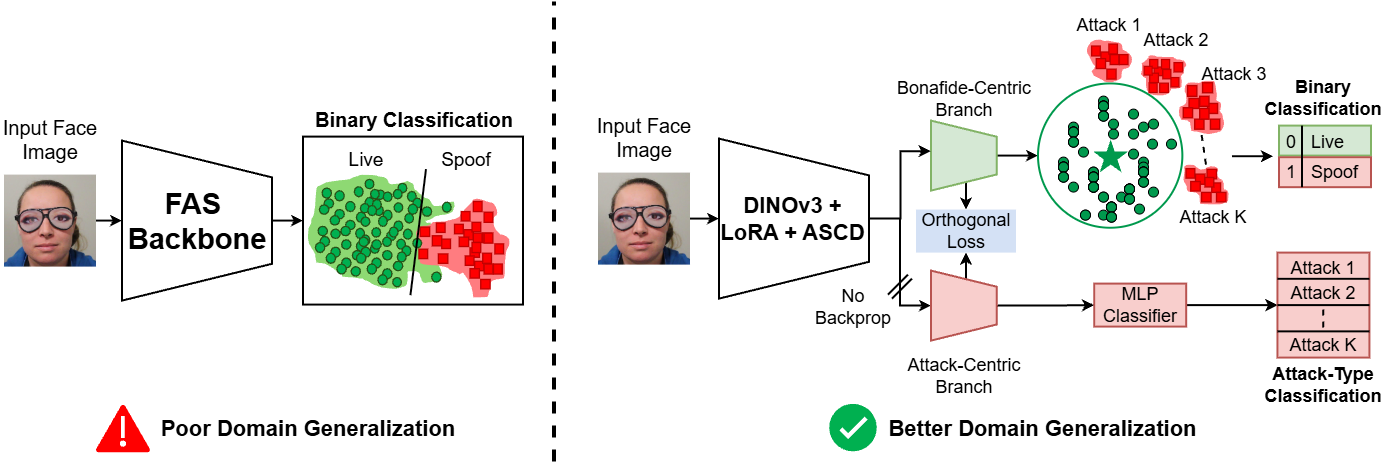}
    \caption{Conceptual overview of DINOFAS. The proposed framework replaces conventional binary classification with bonafide-centric representation learning, where a prototype-based liveness branch and a detached attack-semantic branch jointly learn robust and complementary representations.}
    \label{fig:DINOFAS-overview}
\end{figure*}

Deep learning-based FAS methods achieve strong performance when training and testing data originate from the same distribution. However, their performance degrades considerably under cross-dataset evaluation due to variations in acquisition devices, illumination, image quality, presentation attack instruments, and environmental conditions \cite{boulkenafet2017oulu, abdullakutty2022deep}. Consequently, many methods learn dataset-specific spoof artifacts rather than intrinsic liveness characteristics, limiting their ability to generalize to unseen attacks. Moreover, collecting diverse FAS datasets remains expensive and constrained by privacy concerns, making robust representation learning increasingly important \cite{chen2021dual, boutros2022elasticface}.

Recent advances in foundation models have opened new opportunities for FAS by providing transferable visual representations learned from large-scale image collections. Recent studies have demonstrated the potential of foundation models and vision-language models for improving FAS robustness, while also highlighting the challenge of adapting these representations to domain-specific liveness cues \cite{ozgur2025foundpad, george2021effectiveness, komaty2025exploring}.

Despite their strong representational capability, existing foundation-model-based approaches primarily optimize binary classification objectives, causing the learned representation to become entangled with attack-specific characteristics and dataset-dependent cues. Furthermore, treating all spoof samples as a single negative class overlooks the substantial intra-class variability of presentation attacks, whereas bonafide faces generally exhibit a more stable distribution across domains.

To address these limitations, we propose DINOFAS, a DINOv3-based framework that formulates FAS as a bonafide-centric representation learning problem (Fig.~\ref{fig:DINOFAS-overview}). The proposed framework combines parameter-efficient adaptation of DINOv3 using LoRA, an attention atyle-content disentanglement (ASCD) module for suppressing domain-specific appearance variations, and a dual-branch architecture comprising a prototype-based binary branch and a detached attack-semantic branch. Together, these components encourage the backbone to learn robust liveness representations while preventing attack-specific supervision from contaminating the shared feature space.

The main contributions of this work are summarized as follows:

\begin{itemize}
\item A DINOv3-based face anti-spoofing framework, termed DINOFAS, that formulates FAS as a bonafide-centric representation learning problem.

\item An attention style-content disentanglement (ASCD) module that suppresses domain-specific appearance variations through gated style-content decomposition and label-aware style borrowing.

\item A dual-branch learning framework comprising a prototype-based binary branch and a detached attack-semantic branch, enabling complementary representation learning while preventing attack-category supervision from contaminating the shared backbone representation.

\item Comprehensive experimental evaluation on multiple public FAS benchmarks, demonstrating the effectiveness of the proposed framework under cross-dataset, synthetic-to-real, and protocol-based evaluation settings.
\end{itemize}


\section{Related Works}

\subsection{Cross-Domain Face Anti-Spoofing}

Face anti-spoofing (FAS) has evolved from handcrafted feature descriptors based on texture, frequency analysis, and motion cues \cite{chingovska2012effectiveness, galbally2014biometric, hasan2019face, wang2023learnable, gonzalez2021generalisation} to deep learning methods based on convolutional neural networks (CNNs) \cite{fang2022patchswap, fang2024face, deb2020look} and, more recently, vision transformers \cite{george2021effectiveness, ozgur2025foundpad}. Although these methods achieve strong performance under intra-dataset evaluation \cite{fang2022learnable, liu2018learning, srivatsan2023flip, yu2020fas}, their performance degrades considerably under cross-dataset evaluation due to variations in acquisition devices, illumination, image quality, environments, and presentation attack instruments \cite{fang2024face, shao2020regularized, wang2022face, yan2022domain}.

To address this problem, existing methods primarily adopt domain adaptation (DA) or domain generalization (DG). DA methods align source and target distributions through feature alignment and adversarial learning \cite{li2018unsupervised,wang2019improving}, whereas
DG methods employ adversarial optimization and meta-learning to improve generalization to unseen domains \cite{shao2019multi, jia2020single, chen2021generalizable, shao2020regularized}. More recently, the work in~\cite{kong2024dtda} proposed dual-teacher knowledge distillation with domain alignment (DTDA), combining domain adversarial
perturbations with perceptual and generative teacher models to transfer facial priors and reduce domain discrepancies. Although these approaches improve cross-dataset performance, they often require multiple source domains, additional teacher networks, or specialized optimization strategies \cite{liu2022causal, fang2024face, kong2024dtda}.


Despite these advances, most existing FAS methods continue to formulate spoof detection as a binary classification problem, resulting in representations that become entangled with attack-specific artifacts and dataset-dependent cues, ultimately limiting their generalization to unseen presentation attacks and acquisition conditions.

\subsection{Foundation Models for FAS}

Foundation models (FMs) pretrained on large-scale image collections have demonstrated strong transferability across diverse computer vision tasks through self-supervised and vision-language pretraining \cite{kirillov2023segment, radford2021learning, awais2025foundation}. Representative vision foundation models include SAM \cite{kirillov2023segment}, DINOv2 \cite{oquab2023dinov2}, CLIP \cite{radford2021learning}, and ALIGN \cite{jia2021scaling}. To efficiently adapt these large models to downstream tasks, parameter-efficient fine-tuning methods such as AdaptFormer \cite{chen2022adaptformer}, ViT-Adapter \cite{chen2022vision}, and Low-Rank Adaptation (LoRA) \cite{hu2022lora} have been proposed.

The application of foundation models to FAS remains relatively recent. FoundPAD \cite{ozgur2025foundpad} demonstrated that adapting CLIP with LoRA significantly improves cross-domain FAS performance, while \cite{gonzalez2025foundation} explored the zero-shot capabilities of multiple foundation models for detecting unseen presentation attacks. Although these studies highlight the potential of foundation models for FAS, they primarily formulate spoof detection as a conventional binary classification problem without explicitly disentangling liveness information from attack-specific semantics.

\subsection{Domain-Invariant Representation Learning}

Learning domain-invariant representations is fundamental to improving the robustness of FAS under unseen acquisition conditions and presentation attack instruments. Existing methods primarily achieve this through DA and DG, employing techniques such as feature alignment, adversarial optimization, and meta-learning to reduce domain discrepancies \cite{li2018unsupervised, wang2019improving, shao2019multi, jia2020single, chen2021generalizable, shao2020regularized, fang2024face, wang2024multi}. Although these approaches improve cross-domain performance, they often require multiple source domains, access to target-domain samples, or computationally expensive optimization strategies \cite{liu2022causal, chen2021generalizable}. Methods such as DTDA~\cite{kong2024dtda} address domain discrepancies through adversarial perturbations and auxiliary knowledge transfer. In contrast, DINOFAS emphasizes bonafide-centric prototype learning, attention-based style-content disentanglement, and gradient-decoupled attack supervision without
requiring separate teacher networks.



\section{Methodology}
\label{sec:methodology}

The proposed framework models live faces as a compact reference manifold and treats spoof samples as deviations from this manifold. The framework comprises three main components, a DINOv3 vision transformer~\cite{simeoni2025dinov3} adapted with low-rank adaptation (LoRA) \cite{hu2021lora}, an attention style-content disentanglement (ASCD) module for suppressing domain-specific appearance variations, and a dual-branch learning strategy that decouples bonafide-centric liveness learning from attack-semantic supervision. Given an input face image, the backbone extracts global and spatial representations that are processed by two task-specific branches: a prototype-based binary branch for liveness estimation and a detached attack branch for spoof-category modeling. This design encourages the backbone to learn robust liveness representations while preventing attack-specific gradients from contaminating the shared feature space.

\begin{figure*}[!htbp]
    \centering
    \includegraphics[width=0.95\linewidth]{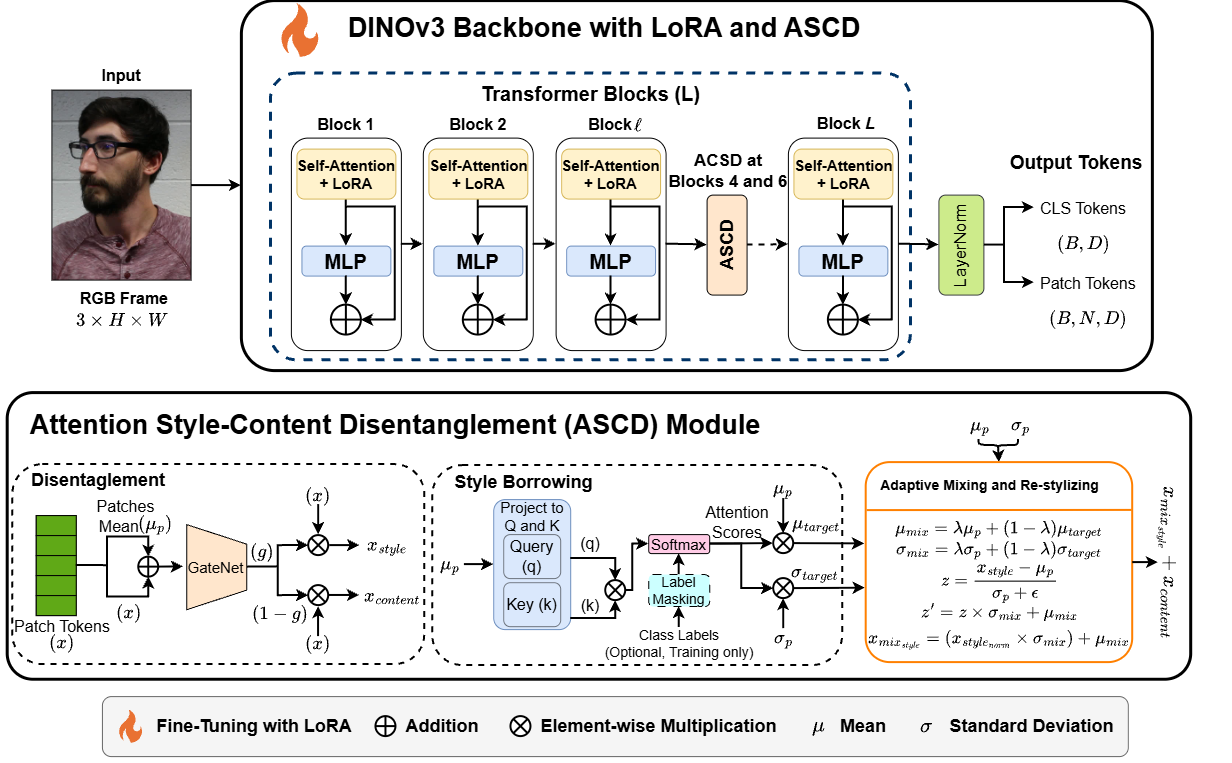}
    \caption{Modified DINOv3 transformer with LoRA adaptation and the proposed attention style-content disentanglement (ASCD) module. The lower panel illustrates the ASCD pipeline, comprising gated style-content decomposition, label-aware style borrowing, and adaptive feature recombination.}
    \label{fig:ascd}
\end{figure*}

\subsection{DINOv3 Backbone with Low-Rank Adaptation}

Let $x \in \mathbb{R}^{3 \times H \times W}$ denote an input RGB face image. A pretrained DINOv3 vision transformer is used as the feature extractor, producing a global class-token representation $z_{\mathrm{cls}}$ and a sequence of spatial patch-token representations $Z_{\mathrm{patch}}$,
\begin{equation}
    z_{\mathrm{cls}}, Z_{\mathrm{patch}} = f_{\theta}(x)
\end{equation}
where $z_{\mathrm{cls}} \in \mathbb{R}^{D}$ and $Z_{\mathrm{patch}} \in \mathbb{R}^{N \times D}$.

To preserve the pretrained visual knowledge, most DINOv3 parameters remain frozen during training. Task-specific adaptation is performed using Low-Rank Adaptation (LoRA), where the attention projection matrix is updated as $W' = W + BA$, with $A$ and $B$ denoting trainable low-rank matrices. Layer normalization parameters are also optimized to allow mild distributional adaptation. As illustrated in Fig.~\ref{fig:ascd}, LoRA adapters are inserted into the transformer attention blocks, while the proposed ASCD module is integrated into the feature extraction pipeline to improve robustness against domain-specific appearance variations.

\subsection{Attention Style-Content Disentanglement}

Face anti-spoofing is highly sensitive to domain-specific appearance variations arising from differences in imaging devices, illumination, compression, and acquisition conditions. To improve robustness, the proposed ASCD module, illustrated in Fig.~\ref{fig:ascd}, decomposes transformer token representations into style and content components, performs label-aware style borrowing, and adaptively recombines the resulting features.

Given a token sequence $T$, a lightweight gating network predicts a soft attention mask
\begin{equation}
    G=\sigma\!\left(\phi(T+\bar{T}_{\mathrm{patch}})\right)
\end{equation}
where $\bar{T}_{\mathrm{patch}}$ denotes the mean patch-token context. The token representation is decomposed into style and content components as $T_{\mathrm{style}}=G\odot T$ and $T_{\mathrm{content}}=(1-G)\odot T$. To preserve global semantic information, style statistics are computed only from the patch tokens, resulting in the patch-wise mean $\mu$ and standard deviation $\sigma_s$.

The ASCD module performs attention style borrowing across samples within the mini-batch. During training, style exchange is restricted to samples from the same class to avoid unrealistic mixing between live and spoof distributions. Let $(\mu_t,\sigma_t)$ denote the style statistics borrowed through the attention mechanism. The source and borrowed statistics are adaptively fused as
\begin{equation}
\mu_{\mathrm{mix}}=\lambda\mu+(1-\lambda)\mu_t,\qquad
\sigma_{\mathrm{mix}}=\lambda\sigma_s+(1-\lambda)\sigma_t
\end{equation}
where $\lambda$ is a learnable mixing coefficient. The mixed style representation is reconstructed by adaptive normalization,
\begin{equation}
T_{\mathrm{style}}'
=
\frac{T_{\mathrm{style}}-\mu}{\sigma_s}
\odot
\sigma_{\mathrm{mix}}
+
\mu_{\mathrm{mix}},
\end{equation}
and recombined with the preserved content representation,
\begin{equation}
T'
=
T_{\mathrm{style}}'
+
T_{\mathrm{content}}
\end{equation}
This regularization encourages the backbone to suppress domain-specific style variations while preserving liveness-related content.

\subsection{Spatial Projection for Localized Attack Evidence}

\begin{figure*}[]
    \centering
    \includegraphics[width=0.75\linewidth]{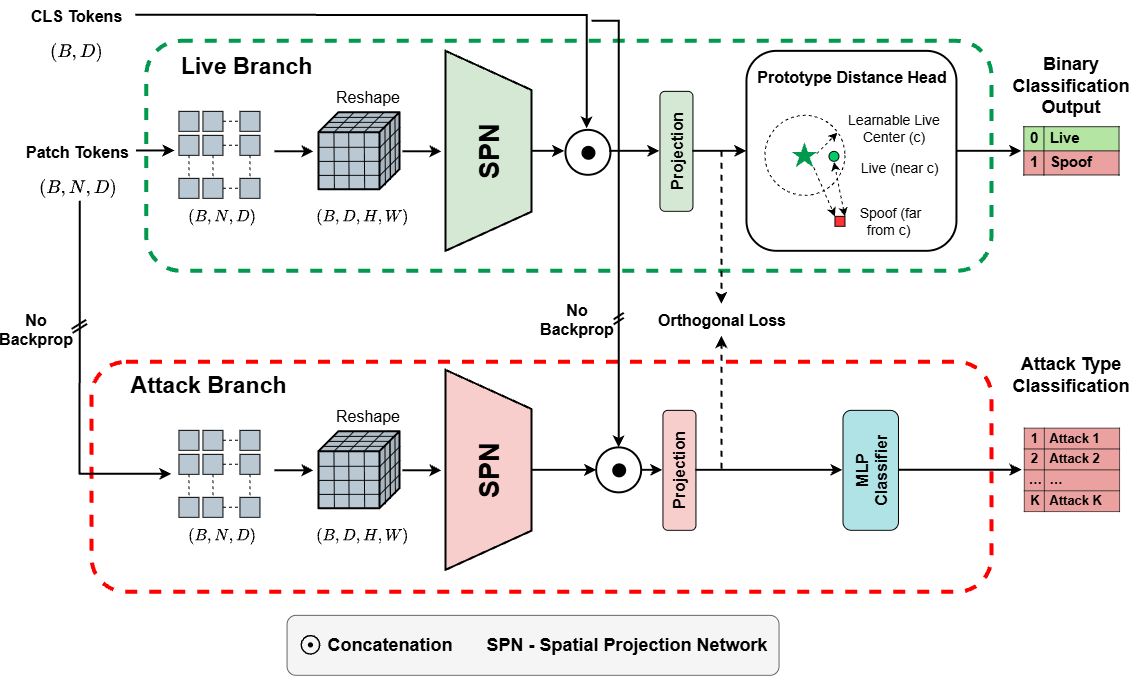}
    \caption{Task-specific branches of the proposed framework. The upper branch performs bonafide-centric prototype learning for binary liveness estimation, while the lower branch learns attack semantics through a detached classifier to prevent attack-specific supervision from contaminating the shared representation.}
    \label{fig:branches}
\end{figure*}

Presentation attacks often exhibit localized artifacts such as mask boundaries, print texture, screen reflections, and partial occlusions. Relying solely on the global class token may therefore discard useful spatial evidence. To incorporate both global and local information, each branch employs a lightweight spatial projection network (SPN), illustrated in Fig.~\ref{fig:branches}.

The patch tokens obtained from the DINOv3 backbone are reshaped into a two-dimensional spatial grid and processed by the SPN using a $3\times3$ depthwise convolution, GroupNorm (up to 32 groups), GELU, spatial dropout ($p=0.2$), a $1\times1$ pointwise convolution, and $1\times1$ adaptive average pooling. The resulting $D$-dimensional local feature is concatenated with the $D$-dimensional class token and projected from $2D$ back to $D$ using a fully-connected layer followed by LayerNorm, ReLU, and dropout ($p=0.1$),

\begin{equation}
z=
\psi\!\left(\left[z_{\mathrm{cls}}; \mathrm{Pool}\left(\mathrm{SPN}(Z_{\mathrm{patch}})\right)
\right]
\right)
\end{equation}
where $\psi(\cdot)$ denotes the final projection layer. The resulting feature combines global facial semantics with localized spoof evidence and serves as the input to both the bonafide-centric binary branch and the detached attack-semantic branch.

\subsection{Bonafide-Centric Binary Branch}

As illustrated in the upper part of Fig.~\ref{fig:branches}, the SPN output is forwarded to a bonafide-centric binary head that models the live-face distribution using a learnable prototype. Instead of employing a conventional linear classifier, the binary score is computed as the Euclidean distance between the normalized feature representation $z_{\mathrm{bin}}$ and the normalized liveness prototype $c$:
\begin{equation}
    d(x)=
    \left\|
    \frac{z_{\mathrm{bin}}}{\|z_{\mathrm{bin}}\|_2}
    -
    \frac{c}{\|c\|_2}
    \right\|_2
\end{equation}
where $c$ denotes the learnable liveness prototype.

Live samples are encouraged to remain close to the prototype, while spoof samples are pushed farther away, effectively treating FAS as a bonafide-centric anomaly detection problem. During inference, the prototype distance serves as the final spoof score, where larger distances indicate a higher likelihood of a presentation attack.

\subsection{Detached Attack-Semantic Branch}

As illustrated in the lower part of Fig.~\ref{fig:branches}, the detached attack-semantic branch learns fine-grained spoof categories from detached backbone features. The shared DINOv3 representations are first detached from the computational graph and then processed by the SPN:
\begin{equation}
    z_{\mathrm{atk}}=
    \psi_{\mathrm{atk}}
    \left(
    \mathrm{stopgrad}(z_{\mathrm{cls}}),
    \mathrm{stopgrad}(Z_{\mathrm{patch}})
    \right)
\end{equation}
where the stop-gradient operation prevents the attack-classification objective from updating the shared backbone. The resulting feature is passed to an MLP classifier that predicts the predefined spoof categories:
\begin{equation}
    \hat{y}_{\mathrm{atk}}=
    h_{\mathrm{atk}}(z_{\mathrm{atk}})
\end{equation}
The attack-classification loss is computed only for spoof samples, while live samples are excluded since they do not belong to any spoof category. This enables the attack branch to learn discriminative attack semantics without contaminating the bonafide-centric representation used for liveness estimation.

\subsection{Training Objective}

The two branches are jointly optimized using three complementary objectives: a bonafide-centric prototype loss, an attack-semantic classification loss applied to detached features, and an orthogonality loss that encourages the two branches to learn complementary representations. Let $y_i\in\{0,1\}$ denote the binary label of the $i$-th sample ($0$: live, $1$: attack), and let $d_i$ denote the prototype distance predicted by the binary branch.

The binary branch employs a margin-based SoftPlus objective that pulls live samples towards the bonafide prototype while pushing spoof samples away:
\begin{equation}
\begin{aligned}
\mathcal{L}_{\mathrm{bin}}
&=
\frac{1}{B}
\sum_{i=1}^{B}
\Big[
(1-y_i)\,
\operatorname{softplus}
\left(
\frac{d_i-m_{\mathrm{live}}}{\tau}
\right)
\\
&\qquad\qquad
+
y_i\alpha_{\mathrm{spoof}}\,
\operatorname{softplus}
\left(
\frac{m_{\mathrm{spoof}}-d_i}{\tau}
\right)
\Big]
\end{aligned}
\end{equation}
where $B$ is the batch size, $m_{\mathrm{live}}=0.5$, $m_{\mathrm{spoof}}=1.6$, $\alpha_{\mathrm{spoof}}=5.0$, and $\tau=0.1$. Compared with a hard-margin formulation, the SoftPlus objective provides smoother optimization around the decision boundary.

The attack branch is supervised using cross-entropy computed only on spoof samples, i.e., $i\in\mathcal{M}_{\mathrm{atk}}=\{i\mid a_i>0\}$, where $a_i$ denotes the attack label ($a_i=0$ for live and $a_i>0$ for the predefined spoof categories available in the training dataset, e.g., print, replay, and mask attacks). Since the classifier predicts only spoof categories, attack labels are shifted by one before computing the loss:

\begin{equation}
\mathcal{L}_{\mathrm{atk}}
=
\frac{1}{|\mathcal{M}_{\mathrm{atk}}|}
\sum_{i\in\mathcal{M}_{\mathrm{atk}}}
\operatorname{CE}
\left(
\hat{y}_{\mathrm{atk},i},
a_i-1
\right)
\end{equation}
Label smoothing with a factor of 0.1 is applied, and the loss is set to zero when a mini-batch contains no spoof samples.

To encourage the two branches to encode complementary information, an orthogonality loss minimizes the absolute cosine similarity between the normalized binary and attack representations:
\begin{equation}
\mathcal{L}_{\mathrm{orth}}
=
\frac{1}{B}
\sum_{i=1}^{B}
\left|
\left\langle
\frac{z_{\mathrm{bin},i}}{\|z_{\mathrm{bin},i}\|_2},
\frac{z_{\mathrm{atk},i}}{\|z_{\mathrm{atk},i}\|_2}
\right\rangle
\right|
\end{equation}

The overall training objective is
\begin{equation}
\mathcal{L}_{\mathrm{total}}
=
\lambda_{\mathrm{bin}}\mathcal{L}_{\mathrm{bin}}
+
\lambda_{\mathrm{atk}}\mathcal{L}_{\mathrm{atk}}
+
\lambda_{\mathrm{orth}}\mathcal{L}_{\mathrm{orth}}
\end{equation}
where $\lambda_{\mathrm{bin}}$, $\lambda_{\mathrm{atk}}$, and $\lambda_{\mathrm{orth}}$ are the loss weights. 

\subsection{Inference}
At inference, only the binary branch is required, the prototype distance $S(x)=d(x)$ is used as the spoof score and thresholded to obtain the live/attack decision. ASCD style borrowing in the DINOFAS backbone is disabled during inference and the module acts as an identity mapping module.


\section{Results}

\begin{table*}[!htb]
\centering
\caption{Single-source cross-dataset evaluation when training on the SynthASpoof dataset. Lower HTER and higher AUC indicate better performance. The best and second-best results are highlighted in \textbf{bold} and \underline{underlined}, respectively.}
\label{tab:synthaspoof}
\scriptsize
\resizebox{\linewidth}{!}{
\begin{tabular}{lcccccccccc}
\toprule
\multirow{2}{*}{\textbf{Method}} &
\multicolumn{2}{c}{\textbf{M}} &
\multicolumn{2}{c}{\textbf{C}} &
\multicolumn{2}{c}{\textbf{I}} &
\multicolumn{2}{c}{\textbf{O}} &
\multicolumn{2}{c}{\textbf{Average}} \\
\cmidrule(lr){2-3}
\cmidrule(lr){4-5}
\cmidrule(lr){6-7}
\cmidrule(lr){8-9}
\cmidrule(lr){10-11}
&
HTER(\%)$\downarrow$ &
AUC(\%)$\uparrow$ &
HTER(\%)$\downarrow$ &
AUC(\%)$\uparrow$ &
HTER(\%)$\downarrow$ &
AUC(\%)$\uparrow$ &
HTER(\%)$\downarrow$ &
AUC(\%)$\uparrow$ &
HTER(\%)$\downarrow$ &
AUC(\%)$\uparrow$\\
\midrule

ResNet~\cite{fang2023synthaspoof} & 25.48 & 79.54 & 39.22 & 62.00 & 8.90 & \underline{96.96} & 34.23 & 71.48 & 26.96 & 77.50\\

PixBiS~\cite{fang2023synthaspoof} & 38.33 & 63.87 & 38.44 & 64.79 & \underline{7.50} & 96.88 & 35.77 & 63.50 & 30.74 & 72.26\\

ViT-SIDE B~\cite{fang2023synfacepad} & 36.67 & 69.78 & 33.33 & 75.21 & 9.80 & 96.67 & \textbf{13.26} & \textbf{94.04} & \underline{23.27} & 83.93\\

SynFace Co-Former A~\cite{fang2023synfacepad} & \underline{18.57} & \underline{90.76} & 41.11 & 64.49 & 16.30 & 92.31 & 21.67 & 86.44 & 24.41 & 83.50\\

SynFace Co-Former B~\cite{fang2023synfacepad} & \textbf{16.67} & \textbf{91.61} & 40.00 & 63.05 & 18.80 & 88.20 & 25.35 & 82.02 & 25.21 & 81.22\\

CoDe-Lc~\cite{fang2023synfacepad} & 37.14 & 71.45 & 37.11 & 69.08 & 12.10 & 95.31 & 37.58 & 66.30 & 30.98 & 75.54\\

CoDe-Lh~\cite{fang2023synfacepad} & 39.05 & 70.58 & 39.33 & 63.70 & 13.90 & 93.84 & 38.11 & 68.33 & 32.60 & 74.11\\

OrthPADNet~\cite{fang2023synfacepad} & 20.95 & 87.59 & 39.78 & 67.32 & 23.70 & 79.55 & 34.92 & 71.69 & 29.84 & 76.54\\

idvcVT~\cite{fang2023synfacepad} & 45.71 & 64.58 & 56.44 & 41.82 & 23.10 & 85.08 & 51.09 & 49.68 & 44.09 & 60.29\\

hdafVPAD~\cite{fang2023synfacepad} & 65.71 & 31.95 & 71.33 & 21.95 & 47.80 & 52.10 & 37.89 & 66.49 & 55.68 & 43.12\\

FoundPAD (ViT-B) \cite{ozgur2025foundpad} & 47.14 & 66.18 & \underline{27.33} & \underline{83.03} & 16.15 & 90.79 & 33.12 & 73.56 & 30.94 & 78.39\\

FoundPAD (ViT-L) \cite{ozgur2025foundpad}
& 45.71 & 69.76 & \textbf{9.89} & \textbf{96.03} & \textbf{6.40} & \textbf{98.58} & 32.05 & 75.69 & 23.51 & \underline{85.01} \\
\midrule
\textbf{DINOFAS (ViT-B) (Proposed)}
& 25.83 & 83.06 &31.30 & 75.54 & 9.12 & 96.57 & \underline{19.40} & \underline{89.39} & \textbf{21.41} & \textbf{86.14} \\

\bottomrule
\end{tabular}}
\end{table*}

\begin{table*}[!htbp]
\centering
\caption{Cross-domain Face Presentation Attack Detection results on the OCIM benchmark. Lower HTER and higher AUC indicate better performance. The best results are shown in \textbf{bold}, while the second-best results are \underline{underlined}.}
\label{tab:cross_domain}
\resizebox{\linewidth}{!}{
\begin{tabular}{lcccccccccc}
\toprule
\multirow{2}{*}{Method} &
\multicolumn{2}{c}{OCI$\rightarrow$M} &
\multicolumn{2}{c}{OMI$\rightarrow$C} &
\multicolumn{2}{c}{OCM$\rightarrow$I} &
\multicolumn{2}{c}{ICM$\rightarrow$O} &
\multirow{2}{*}{Avg. HTER(\%)$\downarrow$} &
\multirow{2}{*}{Avg. AUC(\%)$\uparrow$} \\
\cmidrule(lr){2-3}\cmidrule(lr){4-5}\cmidrule(lr){6-7}\cmidrule(lr){8-9}
& HTER $\downarrow$ & AUC $\uparrow$ & HTER $\downarrow$ & AUC $\uparrow$ & HTER $\downarrow$ & AUC $\uparrow$ & HTER $\downarrow$ & AUC $\uparrow$ & & \\
\midrule
NAS-FAS~\cite{yu2020fas}         & 19.53 & 88.63 & 16.54 & 90.18 & 14.51 & 93.84 & 13.80 & 93.43 & 16.10 & 91.52 \\
LMFD~\cite{fang2022learnable}            & 10.48 & 94.55 & 12.50 & 94.17 & 18.49 & 84.72 & 12.41 & 94.95 & 13.47 & 92.10 \\
MADDG~\cite{shao2019multi}           & 17.69 & 88.06 & 24.50 & 84.51 & 22.19 & 84.99 & 27.89 & 80.02 & 23.07 & 84.40 \\
RFM~\cite{shao2020regularized}             & 17.30 & 90.48 & 13.89 & 93.98 & 20.27 & 88.16 & 16.45 & 91.16 & 16.98 & 90.95 \\
SSDG-R~\cite{jia2020single}          & 7.38  & 97.17 & 10.44 & 95.94 & 11.71 & \underline{96.59} & 15.61 & 91.54 & 11.29 & 95.31 \\
D$^2$AM~\cite{chen2021generalizable}            & 12.70 & 95.66 & 20.98 & 85.58 & 15.43 & 91.22 & 15.27 & 90.87 & 16.10 & 90.83 \\
ViT~\cite{huang2022adaptive}             & \underline{4.75}  & \textbf{98.59} & 15.70 & 92.76 & 17.68 & 86.66 & 16.46 & 90.37 & 13.65 & 92.10 \\
TransFAS~\cite{wang2022face}        & 7.08  & 96.69 & 9.81  & 96.13 & 10.12 & 95.53 & 15.52 & 91.10 & 10.63 & 94.86 \\
CIFAS~\cite{liu2022causal}           & 5.95  & 96.32 & 10.66 & 95.30 & \textbf{8.50}  & \textbf{97.24} & 13.17 & 93.44 & \underline{9.57}  & \underline{95.58} \\

SSAN-M~\cite{wang2022domain} & 10.42 & 94.76 & 16.47 & 90.81 & 14.00 & 96.58 & 19.51 & 88.17 & 15.10 & 92.08 \\

GDA~\cite{zhou2022generative} & 9.20 & 98.00 & 12.20 & 93.00 & 10.00 & 96.00 & 14.40 & 92.60 & 11.45 & 94.90 \\

CF-PAD~\cite{fang2024face}          & 8.11  & 96.43 & 11.78 & 95.64 & 16.50 & 91.50 & 9.87  & 95.13 & 11.57 & 94.68 \\
DTDA~\cite{kong2024dtda}            & 5.71 & 98.03 & 6.67 & 97.27 & 13.12 & 92.24 & 13.13 & 94.24 & 9.66 & 95.45 \\
FoundPAD (ViT-B) \cite{ozgur2025foundpad}    & 20.95 & 89.88 & \textbf{4.89}  & \underline{98.08} & 10.45 & 95.80 & \underline{6.19}  & \underline{98.31} & 10.62 & 95.52 \\
FoundPAD (ViT-L) \cite{ozgur2025foundpad}    & 16.90 & 93.18 & \underline{6.00}  & \textbf{98.72} & \underline{9.90} & 96.07 & \textbf{5.87}  & \textbf{98.41} & 9.67 & \textbf{96.60} \\
\midrule
\textbf{DINOFAS (ViT-B) (Proposed)}   & \textbf{3.75} & \underline{98.39} & 7.59 & 95.99 & 17.25 & 86.05 & 7.77 & 97.16 & \textbf{9.09} & 94.40 \\
\bottomrule
\end{tabular}}
\end{table*}

\subsection{Datasets and Evaluation Protocols}

The proposed method is evaluated on multiple publicly available FAS datasets, including SynthASpoof~\cite{fang2023synthaspoof}, Oulu-NPU (O)~\cite{boulkenafet2017oulu}, CASIA-FASD (C)~\cite{zhang2012face}, Idiap Replay-Attack (I)~\cite{chingovska2012effectiveness}, MSU-MFSD (M)~\cite{wen2015face}, and SiW-Mv2~\cite{guo2022multi}. Cross-dataset generalization is first assessed under a challenging synthetic-to-real protocol, where the model is trained exclusively on the SynthASpoof dataset and evaluated on the M, C, I, and O datasets. The proposed framework is further evaluated on the standard OCIM leave-one-out benchmark, where configurations such as OCI$\rightarrow$M denote training on Oulu-NPU, CASIA-FASD, and Idiap Replay-Attack while testing on MSU-MFSD; OMI$\rightarrow$C, OCM$\rightarrow$I, and ICM$\rightarrow$O follow the same protocol. Finally, robustness to specific sources of domain variation is evaluated using the standard OULU-NPU and SiW-Mv2 protocols, which assess generalization across unseen environmental conditions, presentation attack instruments, acquisition devices, and their combined variations.

\subsection{Implementation Details}

The proposed framework is implemented in PyTorch and trained using the AdamW optimizer for a maximum of 120 epochs with early stopping (patience = 30) and a batch size of 300. LoRA adapters and task-specific heads are optimized using learning rates of $5\times10^{-6}$ and $5\times10^{-4}$, respectively, with a weight decay of $0.05$. Mixed-precision training, gradient clipping ($1.0$), and weighted random sampling are employed throughout training to improve optimization stability and mitigate class imbalance. Input images are resized to $224\times224$ and augmented using random horizontal flipping, color jittering, Gaussian blur, JPEG compression, and Gaussian noise. During evaluation, the decision threshold is determined independently for each evaluation protocol by maximizing Youden's index on the corresponding validation split and is subsequently used to compute HTER, ACER, APCER, and BPCER. Unless otherwise stated, the loss weights are set to $\lambda_{\mathrm{bin}}=1.0$, $\lambda_{\mathrm{attack}}=0.5$, and $\lambda_{\mathrm{orth}}=0.2$.

\subsection{Comparison with State-of-the-Art Methods}

\paragraph{Synthetic-to-Real Generalization}

Herein, we train DINOFAS exclusively on the SynthASpoof dataset and evaluate it on four unseen FAS benchmarks (M, C, I, and O). The results are summarized in Table~\ref{tab:synthaspoof}. DINOFAS achieves the lowest average HTER of 21.41\% and the highest average AUC of 86.14\%, outperforming previous synthetic-data-based FAS methods, including FoundPAD~\cite{ozgur2025foundpad}. Although the proposed framework does not obtain the best performance on every individual target dataset, it consistently maintains competitive results across all evaluation benchmarks, leading to the best overall cross-dataset performance. These results demonstrate that the proposed bonafide-centric representation learning strategy transfers effectively from synthetic training data to previously unseen real-world domains.

\paragraph{Cross-Dataset Evaluation}

Table~\ref{tab:cross_domain} presents the results on the standard OCIM leave-one-out benchmark. DINOFAS achieves the lowest average HTER of 9.09\%, outperforming CIFAS~\cite{liu2022causal} (9.57\%), DTDA~\cite{kong2024dtda} (9.66\%), and FoundPAD (ViT-L)~\cite{ozgur2025foundpad} (9.67\%). The proposed method achieves the lowest HTER on OCI$\rightarrow$M (3.75\%) and competitive performance on OMI$\rightarrow$C (7.59\%) and
ICM$\rightarrow$O (7.77\%). Compared to DTDA, DINOFAS improves HTER on OCI$\rightarrow$M and ICM$\rightarrow$O, but performs worse on OMI$\rightarrow$C and OCM$\rightarrow$I.
The latter also contributes to the lower average AUC of DINOFAS (94.40\%) compared with DTDA (95.45\%). These results indicate that the proposed representation learning strategy improves aggregate HTER without consistently outperforming competing methods on every target domain.

\begin{figure}[!t]
    \centering
    \includegraphics[width=\linewidth]{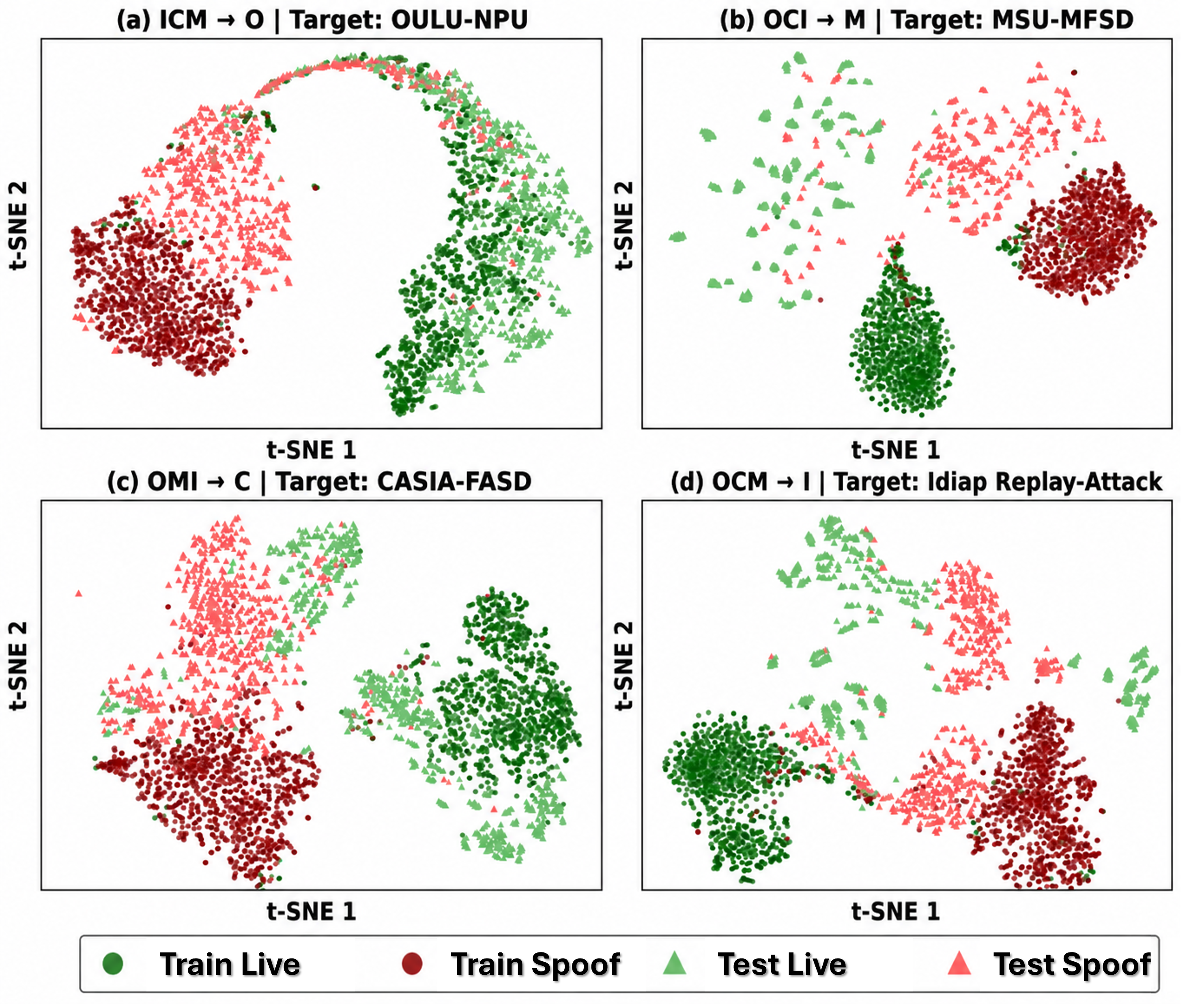}
    \caption{t-SNE visualization of DINOFAS representations under the four OCIM cross-domain protocols.}
    \label{fig:tsne_ocim}
\end{figure}
Figure~\ref{fig:tsne_ocim} visualizes the learned representations across the four OCIM protocols. Bonafide and attack samples remain largely separated across unseen target domains, while the increased overlap for OCM$\rightarrow$I is consistent with its comparatively lower quantitative performance. 

\subsection{Further Generalization Analysis}

To further assess the robustness of the proposed framework under different sources of domain variation, Table~\ref{tab:OULU_protocol_results} reports the results on the four standard evaluation protocols of the OULU-NPU dataset. These protocols evaluate generalization across unseen environmental conditions (Protocol 1), presentation attack instruments (Protocol 2), acquisition devices (Protocol 3), and their combined variations (Protocol 4).

\begin{table}[!htbp]
\centering
\caption{Comparison of Face Presentation Attack Detection performance on the OULU-NPU dataset under the four standard evaluation protocols. Lower ACER, APCER, and BPCER indicate better performance. The best and second-best results for each protocol are highlighted in \textbf{bold} and \underline{underlined}, respectively.}
\label{tab:OULU_protocol_results}
\scriptsize
\resizebox{\columnwidth}{!}{
\begin{tabular}{c|l|ccc}
\toprule
\textbf{P} & \textbf{Method} &
\textbf{ACER} &
\textbf{APCER} &
\textbf{BPCER} \\
\midrule


\multirow{4}{*}{1}
& LMFD-PAD \cite{fang2022learnable} & \underline{1.50} & \underline{1.40} & \underline{1.60} \\
& PatchSwap \cite{fang2022patchswap} & \textbf{0.60} & \textbf{0.40} & \textbf{0.80} \\
& MAX (DINO, CLIP) \cite{gonzalez2025foundation} & 2.64 & 2.50 & 2.77 \\
& \textbf{DINOFAS (Proposed)} & 1.67 & 1.67 & 1.67 \\
\midrule

\multirow{4}{*}{2}
& LMFD-PAD \cite{fang2022learnable}  & 2.00 & 3.10 & \textbf{0.80} \\
& PatchSwap \cite{fang2022patchswap} & \textbf{1.80} & 2.50 & 1.10 \\
& MAX (DINO, CLIP) \cite{gonzalez2025foundation} & 2.13 & \textbf{2.04} & 2.22 \\
& \textbf{DINOFAS (Proposed)} & 1.87 & \underline{2.08} & 1.67 \\
\midrule

\multirow{4}{*}{3}
& LMFD-PAD \cite{fang2022learnable} & 3.40$\pm$3.10 & 3.50$\pm$3.20 & 3.40$\pm$3.10 \\
& PatchSwap \cite{fang2022patchswap} & 3.30$\pm$4.90 & \underline{1.40$\pm$1.30} & 5.30$\pm$10.00 \\
& MAX (DINO, CLIP) \cite{gonzalez2025foundation} & \underline{2.61$\pm$0.66} & 2.63$\pm$0.67 & \underline{2.59$\pm$0.65} \\
& \textbf{DINOFAS (Proposed)} & \textbf{0.17$\pm$0.20} & \textbf{0.34$\pm$0.41} & \textbf{0.00$\pm$0.00} \\
\midrule

\multirow{4}{*}{4}
& LMFD-PAD \cite{fang2022learnable} & \underline{3.30$\pm$3.10} & \textbf{2.50$\pm$4.10} & \underline{3.30$\pm$3.10} \\
& PatchSwap \cite{fang2022patchswap} & 3.80$\pm$6.30 & 2.50$\pm$8.30 & 5.00$\pm$4.20 \\
& MAX (DINO, CLIP) \cite{gonzalez2025foundation} & 3.42$\pm$0.85 & \underline{3.06$\pm$1.12} & 3.78$\pm$0.69 \\
& \textbf{DINOFAS (Proposed)} & \textbf{2.92$\pm$2.33} & 5.00$\pm$5.24 & \textbf{0.83$\pm$2.04} \\
\bottomrule
\end{tabular}
}
\end{table}

On the OULU-NPU benchmark, DINOFAS demonstrates competitive performance across all four evaluation protocols. Under Protocols 1 and 2, which evaluate robustness to unseen environmental conditions and presentation attack instruments, respectively, the proposed framework achieves performance comparable to recent state-of-the-art methods while maintaining low ACER, APCER, and BPCER. The most significant improvement is observed on Protocol 3, which evaluates camera interoperability using the leave-one-camera-out protocol. Here, DINOFAS substantially outperforms the state-of-the-art methods, reducing the ACER from the previous best of $2.61\pm0.66$\% obtained by MAX (DINO, CLIP)~\cite{gonzalez2025foundation} to $\mathbf{0.17\pm0.20}$\%, while also achieving the lowest APCER and a zero BPCER. Under the most challenging Protocol 4, where environmental conditions, presentation attack instruments, and acquisition devices vary simultaneously, DINOFAS achieves the lowest ACER (2.92$\pm$2.33\%) and BPCER (0.83$\pm$2.04\%), demonstrating strong robustness under multiple sources of domain variation despite a comparatively higher APCER.

\begin{table}[!t]
\centering
\caption{HTER (\%) on the SiW-Mv2 leave-one-out protocol. Lower values indicate better performance. The best and second result for each attack type are highlighted in \textbf{bold} and \underline{underlined}, respectively.}
\label{tab:siwmv2_hter}
\scriptsize
\resizebox{\columnwidth}{!}{
\begin{tabular}{l|cccc}
\toprule
\textbf{Attack} &
\textbf{MD-FAS} &
\textbf{Meta} &
\textbf{DINO (ViT-B-14)} &
\textbf{DINOFAS} \\
&
\cite{guo2022multi} &
\cite{muhammad2024domain} &
\cite{gonzalez2025foundation} &
(Proposed) \\
\midrule

FunE.     & 29.50 & \underline{14.80} & \textbf{8.52} & 19.62 \\
PEye      & 2.70 & -- & \textbf{0.19} & \underline{0.38} \\
PMouth    & \underline{1.10} & 10.30 & \textbf{0.19} & \textbf{0.19} \\
PaperG.   & 11.90 & 15.20 & \textbf{0.19} & \underline{5.44} \\
Ob.       & \textbf{1.30} & 45.30 & 18.16 & \underline{7.47} \\
Impers.   & 24.50 & 12.90 & \underline{0.39} & \textbf{0.38} \\
Cosmetic  & \underline{10.90} & 12.10 & 18.88 & \textbf{6.32} \\
HalfM.    & 8.00 & 11.70 & \underline{1.47} & \textbf{0.76} \\
Silicone  & 9.20 & 13.60 & \underline{0.58} & \textbf{0.38} \\
TransM.   & \textbf{0.00} & 24.70 & \underline{1.41} & 3.25 \\
Paper     & \underline{0.60} & 3.20 & \textbf{0.00} & \textbf{0.00} \\
Mann.     & 4.00 & 9.50  & \textbf{0.19} & \underline{0.38} \\
Replay    & 17.90 & \textbf{3.90} & 7.05 & \underline{5.93} \\
Print     & \underline{9.60} & \textbf{1.20} & 13.99 & 9.88 \\
\midrule
\textbf{Avg.$\pm$Std.} &
9.40 $\pm$ 8.80 &
13.70 $\pm$ 10.80 &
\underline{5.09 $\pm$ 7.05} &
\textbf{4.31 $\pm$ 5.50} \\
\bottomrule
\end{tabular}}
\end{table}

To further evaluate generalization to unseen presentation attack instruments, Table~\ref{tab:siwmv2_hter} reports the results on the SiW-Mv2 leave-one-out protocol. DINOFAS achieves the best performance on the \emph{Impers.}, \emph{Cosmetic}, \emph{Half Mask}, and \emph{Silicone} protocols, while matching the best HTER on \emph{PMouth} and \emph{Paper}. It further attains the lowest average HTER of 4.31\%, outperforming all competing methods. Performance remains comparatively weaker on \emph{FunE.}, \emph{TransM.}, and \emph{Print}, indicating that these attack types remain more challenging due to their larger appearance variability.

The OULU-NPU and SiW-Mv2 evaluations collectively demonstrate that DINOFAS maintains robust performance under complementary sources of distribution shift, including acquisition conditions, camera sensors, and previously unseen presentation attack types.

\subsection{Ablation Studies}

\subsubsection{Effect of the Backbone and LoRA Adaptation}

\begin{table*}[!htbp]
\centering
\caption{Effect of foundation-model backbone and LoRA adaptation on the OCIM cross-dataset benchmark. The DINOv2/DINOv3 baseline variants use a conventional linear binary head with BCE, whereas DINOFAS rows use the complete proposed framework. The best and second-best results are highlighted in \textbf{bold} and \underline{underlined}, respectively.}
\label{tab:ablation_backbone}
\footnotesize
\resizebox{\linewidth}{!}{
\begin{tabular}{lcccccccccc}
\hline
\textbf{Method} &
\multicolumn{2}{c}{\textbf{OCI$\rightarrow$M}} &
\multicolumn{2}{c}{\textbf{OMI$\rightarrow$C}} &
\multicolumn{2}{c}{\textbf{OCM$\rightarrow$I}} &
\multicolumn{2}{c}{\textbf{ICM$\rightarrow$O}} &
\multirow{2}{*}{\textbf{Avg. HTER} $\downarrow$} &
\multirow{2}{*}{\textbf{Avg. AUC} $\uparrow$}\\
\cmidrule(lr){2-3}
\cmidrule(lr){4-5}
\cmidrule(lr){6-7}
\cmidrule(lr){8-9}
& HTER $\downarrow$ & AUC $\uparrow$ & HTER $\downarrow$ & AUC $\uparrow$ & HTER $\downarrow$ & AUC $\uparrow$ & HTER $\downarrow$ & AUC $\uparrow$ & &\\
\hline
DINOv2 (ViT-B/16) (Frozen) & 22.50 & 83.22 & 20.55 & 87.55 & 16.75 & 86.95 & 30.72 & 74.83 & 22.63 & 83.14\\
DINOv2 (ViT-B/16) + LoRA & 12.50 & 95.77 & 11.85 & 95.26 & \underline{13.37} & \underline{91.40} & 14.68 & 91.44 & 13.10 & 93.47\\
DINOv3 (ViT-B/16) (Frozen) & 23.75 & 81.02 & 23.14 & 83.62 & 25.62 & 80.38 & 25.97 & 80.53 & 24.62 & 81.39\\
{DINOv3 (ViT-B/16) + LoRA} & 13.33 & 94.22 & 11.85 & 95.15 & \textbf{8.87} & \textbf{94.71} & \underline{9.34} & \underline{95.27} & \underline{10.85} & \textbf{94.84}\\
\hline
DINOFAS (ViT-S/16) & 17.08 & 90.41 & 12.40 & 94.74 & 20.06 & 83.03 & 15.52 & 91.72 & 16.26 & 89.97 \\
DINOFAS (ViT-B/16) (ours) & \textbf{3.75} & \textbf{98.39} & \textbf{7.59} & \underline{95.99} & 17.25 & 86.05 & \textbf{7.77} & \textbf{97.16} & \textbf{9.09} & 94.40\\
\textcolor{black}{DINOFAS (ViT-L/16)} & \underline{9.16} & \underline{97.31} & \underline{7.77} & \textbf{97.58} & 16.00 & 89.48 & 12.84 & 93.82 & 11.44 & \underline{94.55} \\
\hline
\end{tabular}
}
\end{table*}

An ablation study comparing different DINO backbone variants on the OCIM cross-dataset benchmark is presented in Table~\ref{tab:ablation_backbone}. For both DINOv2 and DINOv3, parameter-efficient adaptation using LoRA (rank $r=8$, $\alpha=16$, dropout $=0.05$) substantially improves performance over the corresponding frozen backbones, reducing the average HTER from 22.63\% to 13.10\% for DINOv2 and from 24.62\% to 10.85\% for DINOv3. Although the frozen DINOv2 backbone exhibits slightly better generalization than the frozen DINOv3 backbone, DINOv3 consistently achieves superior performance after adaptation, indicating its stronger capacity for learning task-specific liveness representations. We further investigate the influence of DINOv3 model scale by evaluating DINOFAS with ViT-S/16, ViT-B/16, and ViT-L/16 backbones. Among the three variants, ViT-B/16 achieves the best average HTER (9.09\%), outperforming both the smaller ViT-S/16 and the larger ViT-L/16 models. This suggests that the base-sized backbone provides the most favorable balance between representation capacity and adaptation for the proposed bonafide-centric learning framework.

\subsubsection{Effect of the Proposed Learning Strategy}

\begin{table}[!h]
\centering
\caption{Ablation study of the proposed learning strategy on the OCIM cross-dataset benchmark. ASCD is enabled in all configurations, while $L_{\mathrm{bin}}$, $L_{\mathrm{attack}}$, $L_{\mathrm{orth}}$, and the stop-gradient mechanism are varied to evaluate their contributions.}
\label{tab:ablation_strategy}
\resizebox{\columnwidth}{!}{
\begin{tabular}{cc|ccccc}
\hline
\textbf{Cfg.} &
$L_{\mathrm{bin}}$ &
$L_{\mathrm{attack}}$ &
$L_{\mathrm{orth}}$ &
StopGrad &
Avg. HTER (\%) $\downarrow$ \\
\hline
{A} & \checkmark & $\times$ & $\times$ & -- & 14.16 \\
B & \checkmark & \checkmark & $\times$ & $\times$ & 13.95 \\ \hline
C & \checkmark & $\times$ & \checkmark & \checkmark & 14.03 \\
{D} & \checkmark & \checkmark & $\times$ & \checkmark & {13.45} \\
E & \checkmark & \checkmark & \checkmark & $\times$ & 13.30 \\
\hline
Proposed & \checkmark & \checkmark & \checkmark & \checkmark & \textbf{9.09} \\
\hline
\end{tabular}
}
\end{table}

The contribution of the proposed learning strategy on the OCIM cross-dataset benchmark is evaluated in Table~\ref{tab:ablation_strategy}. Configuration A, which optimizes only the bonafide-centric prototype loss ($L_{\mathrm{bin}}$), establishes a baseline with an average HTER of 14.16\%. 
Configuration B incorporates the attack-semantic classification loss ($L_{\mathrm{attack}}$) without stop-gradient, resulting in a modest improvement over the baseline. Enabling the stop-gradient mechanism in configuration D further reduces the average HTER to 13.45\%, indicating that isolating attack supervision from the shared backbone improves generalization .
Introducing the orthogonality loss alone, as expressed by configuration C, provides only a marginal gain over the baseline, whereas combining all three objectives in configuration E, without stop-gradient achieves an average HTER of 13.30\%. 
The complete DINOFAS framework, which jointly optimizes $L_{\mathrm{bin}}$, $L_{\mathrm{attack}}$, and $L_{\mathrm{orth}}$ with stop-gradient enabled, achieves the best performance with an average HTER of 9.09\%. These results show that, on top of the fixed ASCD-enabled backbone, the attack-semantic objective and orthogonality regularization are most effective when attack supervision is decoupled from the shared backbone through the stop-gradient mechanism.

\subsubsection{Effect of ASCD Design Choices}

\begin{table}[!h]
\centering
\caption{Ablation study of the proposed ASCD module on the OCIM cross-dataset benchmark. Lower HTER and higher AUC indicate better performance.}
\label{tab:ablation_ascd}
\scriptsize
\resizebox{\linewidth}{!}{
\begin{tabular}{llcc}
\toprule
\textbf{Component} & \textbf{Configuration} & \textbf{Avg. HTER (\%)}$\downarrow$ & \textbf{Avg. AUC (\%)}$\uparrow$\\
\midrule
\multirow{2}{*}{Style Borrowing}
& Class-agnostic & 13.36 & 92.00\\
& \textbf{Class-aware (Proposed)} & \textbf{9.09} & \textbf{94.40}\\
\midrule
\multirow{2}{*}{Gating Mechanism}
& Without gating & 12.14 & 93.52\\
& \textbf{With gating (Proposed)} & \textbf{9.09} & \textbf{94.40}\\
\midrule
\multirow{4}{*}{ASCD Placement}
& None & 13.32 & 91.87\\
& Early (Blocks 1,2) & 13.67 & 91.84\\
& \textbf{Middle (Blocks 4,6--Proposed)} & \textbf{9.09} & \textbf{94.40}\\
& Late (Blocks 10,11) & 14.02 & 92.35\\
\bottomrule
\end{tabular}}
\end{table}


The effectiveness of the proposed ASCD module is evaluated through three complementary ablation experiments on the OCIM cross-dataset benchmark, as summarized in Table~\ref{tab:ablation_ascd}. First, replacing the proposed class-aware style borrowing with class-agnostic style mixing degrades the average HTER from 9.09\% to 13.36\%, demonstrating the importance of restricting style exchange within the same liveness class. Second, removing the gating mechanism increases the average HTER to 12.14\%, indicating that adaptive style-content decomposition is more effective than uniform feature mixing. Finally, the placement of ASCD is investigated by inserting the modules at different transformer stages. Applying ASCD to the early transformer blocks (Blocks 1, 2) performs similarly to removing the module altogether, while inserting it into the late transformer blocks (Blocks 10, 11) also degrades performance. In contrast, placing ASCD in the middle semantic blocks (Blocks 4 and 6) achieves the best performance, suggesting that intermediate transformer representations provide the most suitable balance between semantic abstraction and domain-specific style information for style-content disentanglement.

\subsubsection{Sensitivity to the Orthogonality Loss Weight}

\begin{figure}[]
    \centering
    \includegraphics[width=\linewidth]{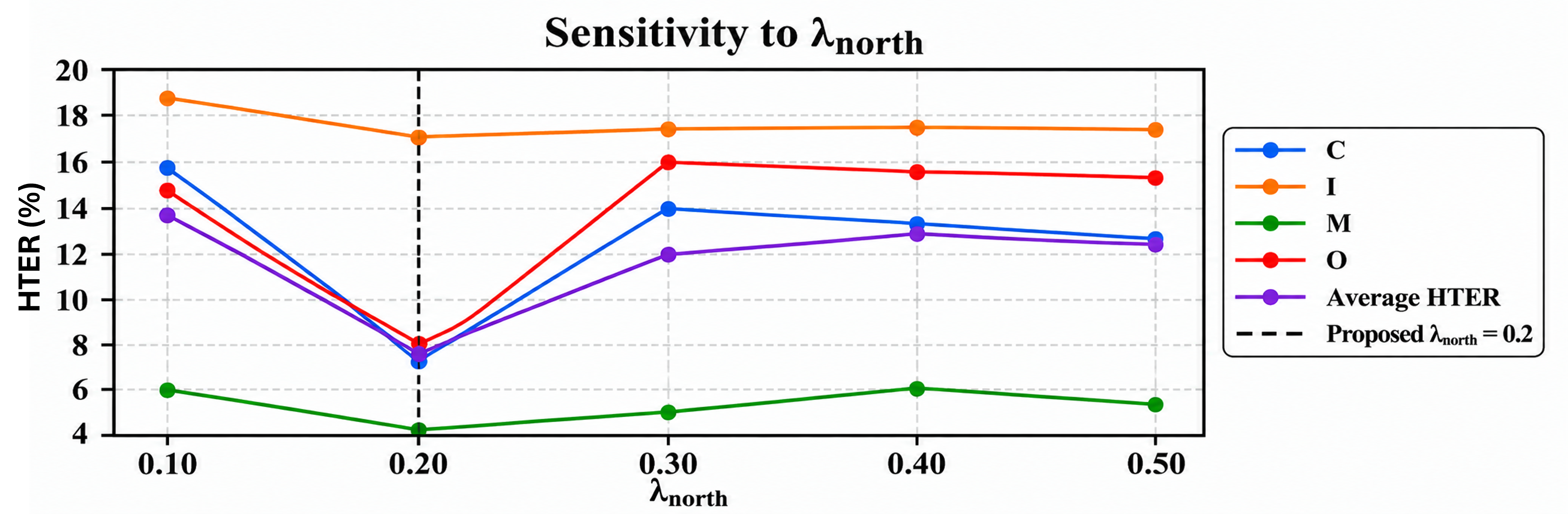}
    \caption{Sensitivity to $\lambda_{\mathrm{orth}}$ on OCIM.}
    \label{fig:orth_weight}
\end{figure}

The influence of the orthogonality loss weight ($\lambda_{\mathrm{orth}}$) is evaluated on the OCIM cross-dataset benchmark by varying its value while keeping all other training settings fixed. As shown in Fig.~\ref{fig:orth_weight}, $\lambda_{\mathrm{orth}}=0.2$ consistently provides the best overall trade-off, achieving the lowest average HTER of 9.09\% while also achieving the best or near-best performance across the individual evaluation protocols. Increasing the orthogonality weight beyond this point degrades performance, particularly on the CASIA-FASD and OULU-NPU datasets, indicating that excessive feature decorrelation can interfere with the primary liveness and attack-semantic learning objectives.

\section{Conclusion}
This paper presented DINOFAS, a DINOv3-based framework for face anti-spoofing that combines bonafide-centric representation learning with parameter-efficient foundation model adaptation. The proposed framework integrates attention style-content disentanglement, detached attack-semantic supervision, and a joint optimization strategy based on bonafide-centric prototype, attack-semantic, and orthogonality objectives to learn robust and complementary liveness representations while reducing reliance on dataset-specific spoof characteristics. Experimental evaluations on multiple public FAS benchmarks demonstrate strong cross-dataset generalization, achieving the best average HTER among the compared methods on both the SynthASpoof-to-real transfer setting and the standard OCIM benchmark, while maintaining robust performance on the OULU-NPU and SiW-Mv2 generalization protocols.






\begin{IEEEbiography}[{\includegraphics[width=1in,height=1.25in,clip,keepaspectratio]{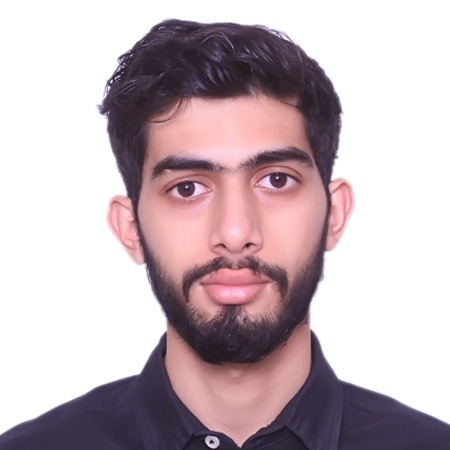}}]{Shakeel Muhammad Ibrahim} received his BS degree in Computer Science from the University of Karachi, Karachi, Pakistan, in 2019, and his MS degree in Computer Engineering from Hallym University, Chuncheon-si, Republic of Korea, in 2025. From 2020 to 2023, he worked as an ML Engineer at Softech Worldwide, focusing on computer vision problems, including object detection, semantic segmentation, and face recognition. During his master's studies, he served as a Graduate Research Assistant at the AI Accelerator Computing Lab at Hallym University. He is currently working as an AI Researcher at Geomexsoft Co., Ltd., in Chuncheon-si, Republic of Korea. His research interests include computer vision and deep learning, with a focus on face recognition, image reconstruction, face presentation attack detection, and aerial segmentation.
\end{IEEEbiography}
\vspace{-3pt}

\begin{IEEEbiography}[{\includegraphics[width=1in,height=1.25in,clip,keepaspectratio]{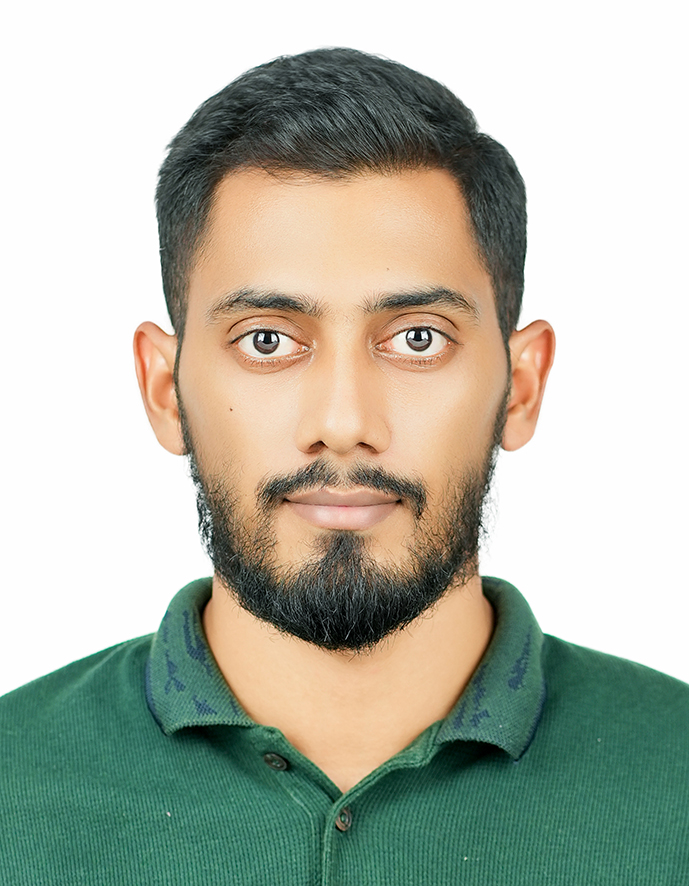}}]{Muhammad Sohail Ibrahim} earned his B.E. degree in Electronic Engineering from Iqra University, Pakistan, in 2012, followed by an M.E. degree in Telecommunications from N.E.D. University of Engineering \& Technology, Pakistan, in 2016, and his Ph.D. degree from Chosun University, Republic of Korea, in 2024. From 2013 to 2019, he served as a lecturer in the Faculty of Engineering, Science, and Technology at Iqra University, Karachi, Pakistan. During 2013-2014, he also worked as a research assistant with the Embedded Systems Research Group at the Karachi Institute of Economics and Technology, Pakistan. During 2024-2025, he conducted his postdoctoral research at Microfluidics and Nano-BioSystems (MNBS) lab. at Kumoh National Institute of Technology, Gumi-si, Republic of Korea. He is currently pursuing his postdoctoral research at The Interdisciplinary Research Center for Intelligent Secure Systems (IRC-ISS) at King Fahd University of Petroleum \& Minerals, Dhahran, Saudi Arabia. His research interests span deep learning, computer vision, bioinformatics, and medical image processing. Muhammad Sohail Ibrahim was honored with the 2020 Highly Cited Review Paper award from Applied Energy, Elsevier, for his paper titled "Machine Learning Driven Smart Electric Systems: Current Trends and New Perspectives".
\end{IEEEbiography}
\vspace{-3pt}

\begin{IEEEbiography}[{\includegraphics[width=1in,height=1.25in,clip,keepaspectratio]{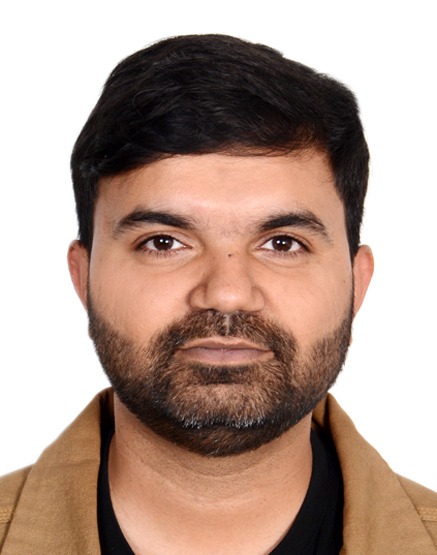}}]{Muhammad Shahid Jabbar} earned his Ph.D. in Computer Engineering from Sungkyunkwan University, Suwon, South Korea, in 2022. He was a Researcher at the H-Lab and the Media System Lab at SKKU. His doctoral research focused on developing modern deep learning solutions using multimodal data for convergence applications involving multi-sensory interaction and cross-modular associations, with particular emphasis on accessibility for visually impaired users. He is currently a Post-Doctoral Fellow at the SDAIA–KFUPM Joint Research Center for Artificial Intelligence, King Fahd University of Petroleum \& Minerals (KFUPM), Dhahran, KSA. Prior to joining KFUPM, he worked as Senior AI Researcher at Radisen Co. Ltd., Seoul, South Korea, where he contributed to AI-powered digital radiography solutions for medical image analysis. His research interests include computer vision, diffusion models, multimodal learning, and inverse problems, with a focus on biomedical and human-centered artificial intelligence.
\end{IEEEbiography}

\vspace{-3pt}

\begin{IEEEbiography}[{\includegraphics[width=1in,height=1.25in,clip,keepaspectratio]{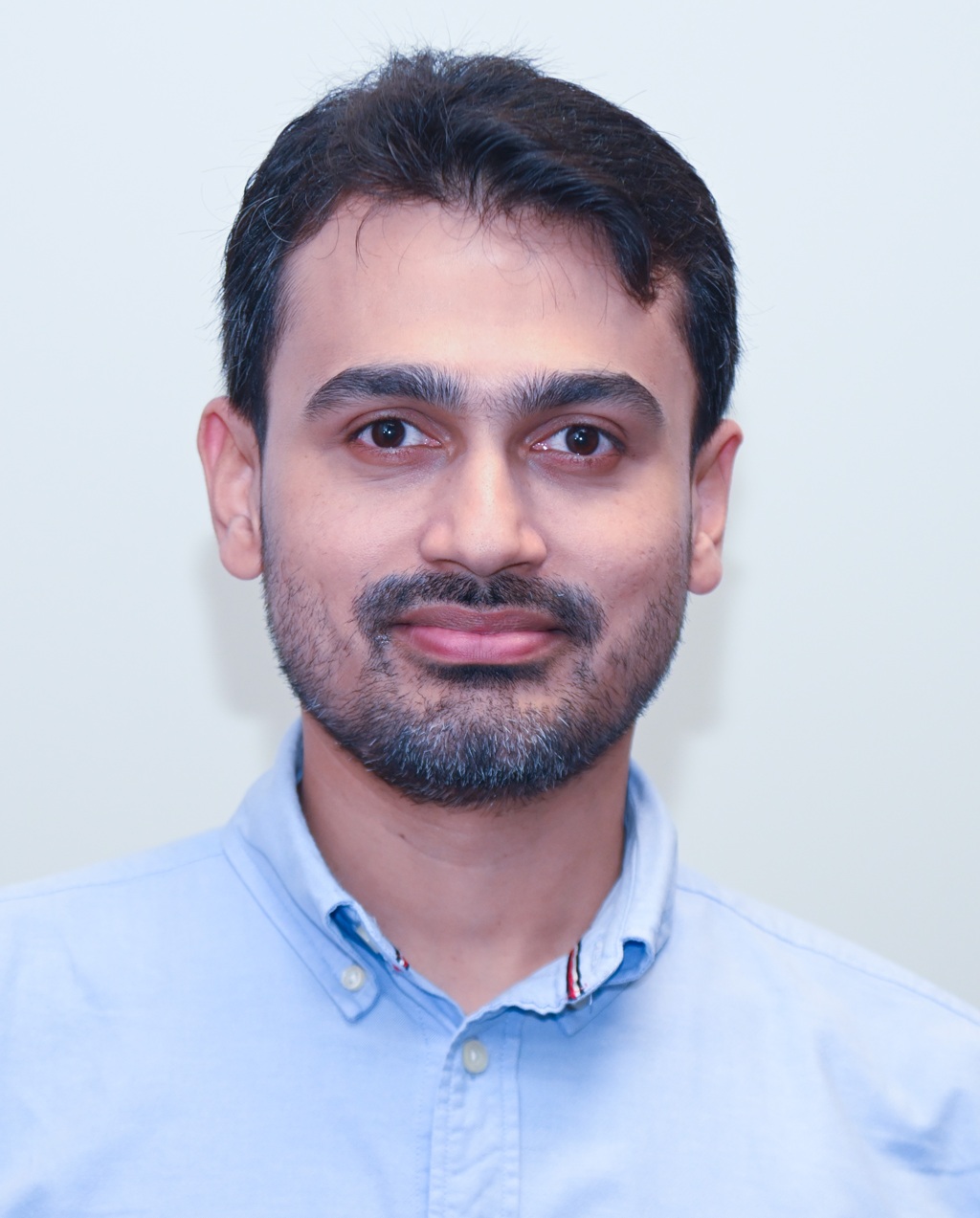}}]{Shujaat Khan} is an Assistant Professor in the Department of Computer Engineering and a Fellow at the Saudi Data and AI Authority (SDAIA) and King Fahd University of Petroleum \& Minerals (KFUPM) under the SDAIA-KFUPM Joint Research Center for Artificial Intelligence (JRC-AI) at KFUPM, Dhahran, KSA. Prior to joining KFUPM, he was a Senior AI Scientist at Digital Technology \& Innovation, Siemens Medical Solutions USA, Inc. He earned his Ph.D. from the Department of Bio and Brain Engineering at the Korea Advanced Institute of Science and Technology (KAIST), Daejeon, South Korea, in 2022. He was a researcher with Synergistic Bioinformatics (SynBi) and the Bio Imaging, Signal Processing Learning (BISPL) Labs at KAIST. His research interests include machine learning, optimization, inverse problems, and signal processing, with a focus on biomedical and bioinformatics applications.
\end{IEEEbiography}

\newpage

 




\vfill

\end{document}